\documentclass[conference]{IEEEtran}
\IEEEoverridecommandlockouts
\usepackage{cite}
\usepackage{amsmath,amssymb,amsfonts}
\usepackage{algorithmic}
\usepackage{graphicx}
\usepackage{textcomp}
\usepackage{xcolor}
\usepackage{svg}
\usepackage{amsmath}
\def\BibTeX{{\rm B\kern-.05em{\sc i\kern-.025em b}\kern-.08em
    T\kern-.1667em\lower.7ex\hbox{E}\kern-.125emX}}
\makeatletter
\def\ps@IEEEtitlepagestyle{
\def\@evenfoot{}
}

\makeatother    
\begin{document}

\title{Collision-Free Navigation using Evolutionary Symmetrical Neural Networks}

\author{\IEEEauthorblockN{
Hesham M. Eraqi}
\IEEEauthorblockA{\textit{Department of Computer Science} \\
\textit{The American University in Cairo}\\
New Cairo, Egypt \\
heraqi@aucegypt.edu}
\and
\IEEEauthorblockN{
Mena NAGIUB}
\IEEEauthorblockA{
\textit{Valeo Schalter und Sensoren GmbH}\\
Bietigheim-Bissingen, Germany \\
mena.nagiub@valeo.com}
\and
\IEEEauthorblockN{Peter Sidra}
\IEEEauthorblockA{\textit{Department of Computer Science} \\
\textit{The American University in Cairo}\\
New Cairo, Egypt \\
peter.sidra.18@ucl.ac.uk}
}

\maketitle

\begin{abstract}
Collision avoidance systems play a vital role in reducing the number of vehicle accidents and saving human lives. This paper extends the previous work using evolutionary neural networks for reactive collision avoidance. We are proposing a new method we have called symmetric neural networks. The method improves the model's performance by enforcing constraints between the network weights which reduces the model optimization search space and hence, learns more accurate control of the vehicle steering for improved maneuvering. The training and validation processes are carried out using a simulation environment - the codebase is publicly available. Extensive experiments are conducted to analyze the proposed method and evaluate its performance. The method is tested in several simulated driving scenarios. In addition, we have analyzed the effect of the rangefinder sensor resolution and noise on the overall goal of reactive collision avoidance. Finally, we have tested the generalization of the proposed method. The results are encouraging; the proposed method has improved the model's learning curve for training scenarios and generalization to the new test scenarios. Using constrained weights has significantly improved the number of generations required for the Genetic Algorithm optimization. \\
\end{abstract}


\section{Introduction}
Designing a control software for a self-driving vehicle is a complex task. The software should concurrently tolerate many scenarios and exceptional cases. It also should maintain and meet reasonable software complexity and resource constraints.

Collision avoidance is a feature that allows a vehicle to move without colliding with the environment. Vehicles can be any robotic system like autonomous cars, unmanned vehicles, etc.\cite{b4}. Reactive collision avoidance controls the motion of the vehicle directly based on the current sensor data to react to unforeseen changes in unknown and dynamic environments. The environment dynamics do not cooperate with the ego-vehicle (i.e., the vehicle that learns) to achieve collision avoidance. Therefore, reactive collision avoidance has a good performance in real-time maneuvering through such scenarios\cite{b3}.
Neural networks are good candidates for controlling vehicles for collision avoidance, especially in dynamic environments, due to their generalization capabilities. They can generalize to many scenarios. However, this generalization comes at the cost of the complexity of initial weights selection and training of the networks using many possible examples. Therefore, generalization is usually achieved using complex training algorithms. Evolutionary algorithms can be an excellent alternative to the selection of the weights and the training of the network\cite{b8}.

\section{Related Work}
An overview of the literature is provided in \cite{b7} for combining deep Artificial Neural Networks (ANN) and Evolutionary Algorithms, specifically Genetic Algorithms (GA), drawing out the common themes and the emerging knowledge about what seems to work and what does not. Multi-layer neural networks, like deep Convolutional Neural Networks (CNN), possess several properties which make them particularly suited to complex pattern classification problems. As for the evolutionary algorithms, they are well suited to the problem of training feed-forward networks as they are good at exploring a large and complex space in an intelligent way to find values close to the global optimum\cite{b6}. Furthermore, evolutionary algorithms can be used to create neural network controllers for simulated cars\cite{b9}. They have evolved controllers with robust performance over different tracks and can work better on particular tracks. Finally, a flexible method for solving the traveling salesman problem using genetic algorithms has been introduced as they can be used to train neural networks producing evolutionary artificial neural networks\cite{b5}.

A new evolutionary-based algorithm has been developed to simultaneously evolve the topology and the connection weights of ANNs through a new combination of Grammatical Evolution (GE) and GA\cite{b1}. GE is adopted to design the network topology while GA is incorporated for better weight adaptation.\\
In our previous paper\cite{b11} we have introduced a new method for vehicle reactive collision avoidance using Evolutionary Neural Networks (ENN). A single front-facing rangefinder sensor is the only input required by the method. The sensor provides the neural network with spatial proximity readings measured at multiple horizontal angles. The neural network learns how to control the vehicle steering wheel angle by directing the vehicle such that it does not collide with the dynamic environment. The neural network guides the vehicle around the environment, and a genetic algorithm is used to pick and breed generations of more intelligent vehicles.\\

In this paper, we are extending the work done in\cite{b11} through the introduction of the new method of the symmetric neural networks. This new method constrains the relation between the neurons' weights by enforcing mathematical odd-symmetry for the weights of each neuron. The expectation is that the method will enhance obstacle detection performance and improve the robustness against sensor noise. The training process, analysis, and validation are carried out using simulation. We have conducted 17 experiments to validate the proposed method analyze and evaluate its performance. The results are encouraging; the proposed method has successfully allowed vehicles to learn collision avoidance in different scenarios that are unseen during training. The scenarios include a vehicle that learns how to navigate safely (i.e., without collision) through a free static track and tracks with obstacles. In addition, we have simulated noise for some sensors to evaluate the robustness of the method and its generalization capabilities.

\section{System Overview}
ANN can be considered an optimization problem in looking for the best weights to achieve some task. GA\cite{b10} is considered as a search algorithm for solution-optimization, which makes GA capable of training a neural network \cite{b7}. ENN, Neuroevolution, or neuro-evolution, is a form of machine learning that uses evolutionary algorithms to train ANN, in other words, estimating the weights of the neural network. It is most commonly applied in artificial life, and intelligent computer games and hence has potential contributions towards self-driving vehicles. 

The chromosome format is chosen to be the vector of real numbers with a sequence of all of the neural network weights. The sequence is sorted layer by layer. The weights of each layer are sorted such that all of the weights going into a neuron are consecutive. The activation function used is an odd version of the sigmoid function, Eq. \ref{eq:sigmoid}, where the final value is calculated as

\begin{equation}
\label{eq:sigmoid}
\sigma \left (x\right )= sigmoid\left (x\right ) - 0.5
\end{equation}

\begin{figure}[htbp]
\centerline{\includegraphics[scale=0.25]{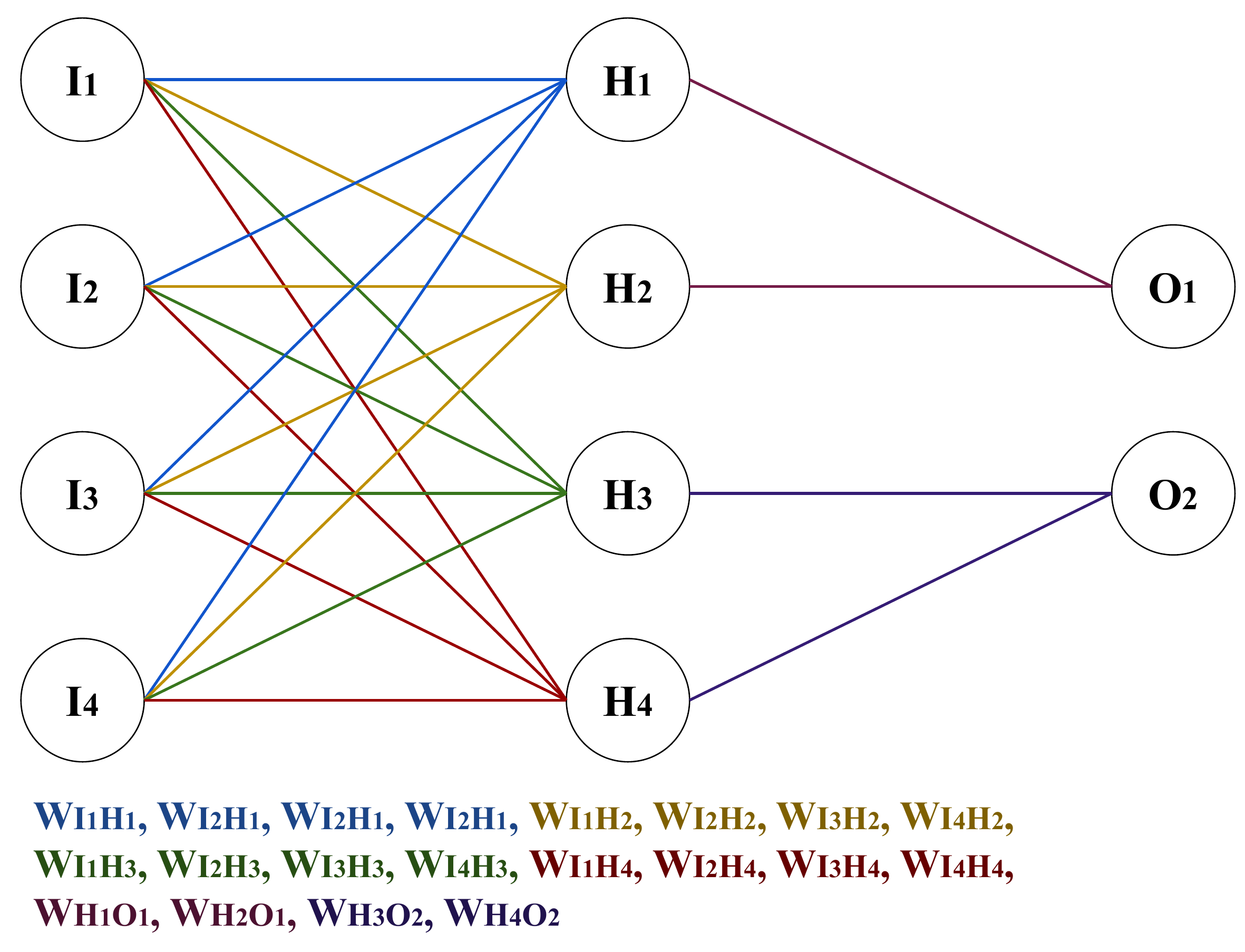}}
\caption{Example of a \begin{math}4 \times 4 \times 2 \end{math} neural network and its chromosome genes representation.}
\label{fig1}
\end{figure}

Figure \ref{fig1} shows an example for a \begin{math}4 \times 4 \times 2 \end{math} neural network and its chromosome.

\subsection{Symmetric Neural Networks}
Searching for a fitting neural network model using GA can be challenging, especially for large networks like deep neural networks. However, by taking into account that inputs of the neurons represent the distance measurement along with the different directions of the rangefinder reflections, a spatial symmetry can be enforced between the weights assigned to input which could improve the network's performance. 

Symmetry can be enforced by giving each input an equal chance of affecting the neuron's activation, improving navigation performance, especially in challenging cases, like corners and intersections, and also in the case of noisy inputs. Constraining the weights through symmetry also helps reduce the search space for the fitting chromosome since the overall chromosome size would be shortened to almost half since half of the chromosome genes are already predefined through enforcing symmetry. 

To study this method a network has been designed to test the inversion of the steering angle output \begin{math}O_{\text{steer}} = -O_{\text{steer\_swapped}}\end{math} when the inputs are swapped i.e. \begin{math}I_{1},I_{2},...I_{n} \rightarrow I_{n},I_{n-1},...I_{1} \end{math}. This behavior has been achieved by constraining the hidden layers links' weights such that for each neuron in the hidden layers and output layer, the first half of its set of weights is the additive inverse of the second half of its set of weights. We have called such constrained network design symmetric neural networks.

\subsection{Symmetric Neural Networks Mathematical Analysis}
\begin{figure}[htbp]
\centerline{\includegraphics[scale=0.25]{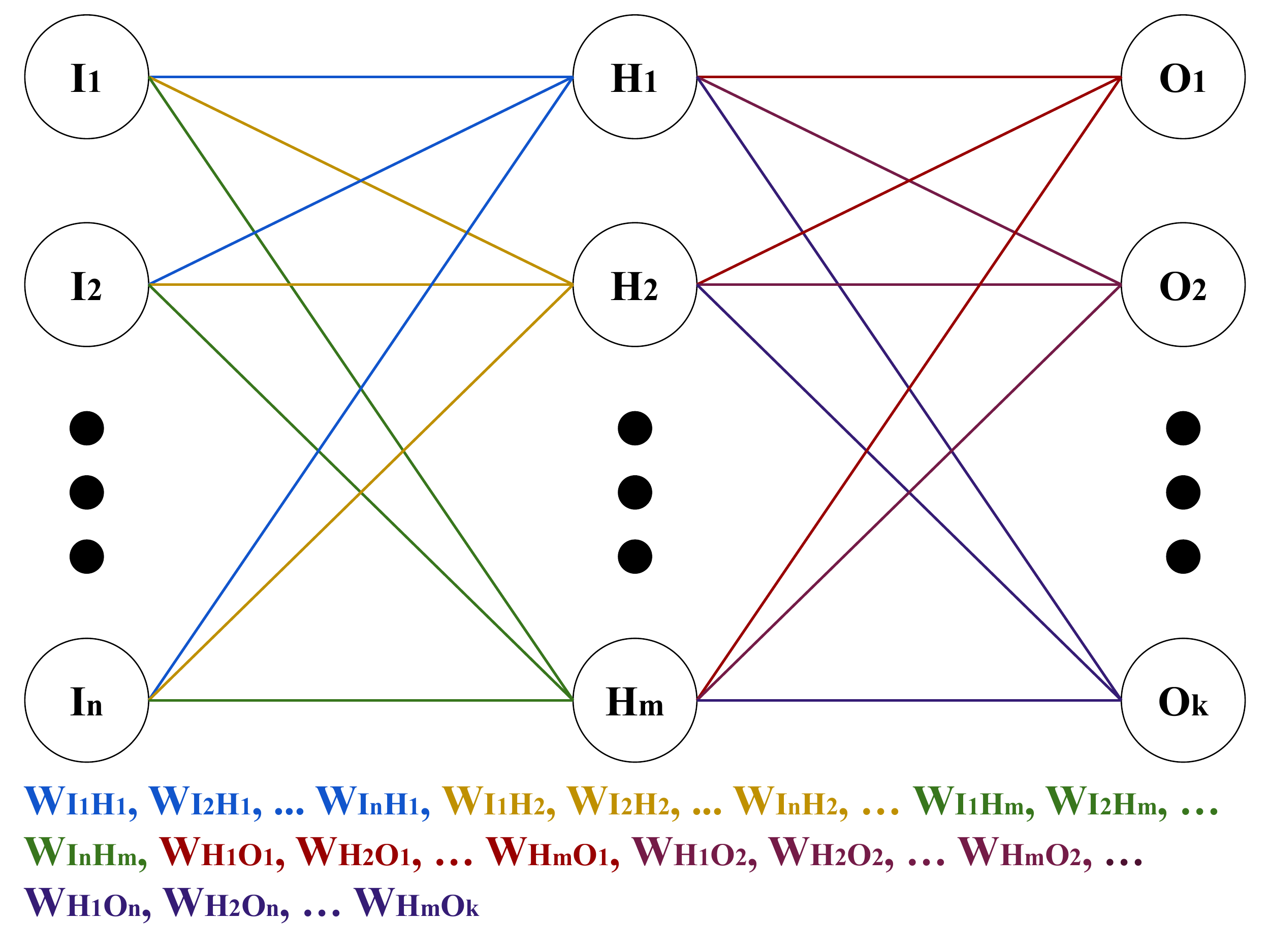}}
\caption{Generic Neural Network Architecture.}
\label{fig2}
\end{figure}

This mathematical analysis illustrates the concept of symmetric neural networks. Considering the network shown in Figure \ref{fig2}, where \begin{math}F(x)\end{math} defines the network output.\\

\resizebox{.9\hsize}{!}{\begin{math}Out_{x} = F\left ( \sum_{i=1}^{m}W_{H_iOut_{x}} \cdot F \left ( \sum_{j=1}^{\frac{n}{2}}W_{I_jH_i} \cdot I_j + 
W_{I_{(n+1-j)}H_i} \cdot I_{n+1-j}\right )\right ) \end{math}}\\

\resizebox{0.5\hsize}{!}{\begin{math}\because W_{I_{(n+1-j)}H_i} = -W_{I_jH_i}, \forall j \in \{1 \rightarrow \frac{n}{2}\}\end{math}}
\begin{equation}
\label{eq:eq1}
\resizebox{.9\hsize}{!}{\begin{math}\therefore Out_{x} = F\left ( \sum_{i=1}^{m}W_{H_iOut_{x}} \cdot F \left ( \sum_{j=1}^{\frac{n}{2}}W_{I_jH_i} \cdot I_j - W_{I_jH_i} \cdot I_{n+1-j}\right )\right )\end{math}}
\end{equation}

\resizebox{0.7\hsize}{!}{\begin{math}{\text{When\;inputs\;are\;swapped\;}}I_{1},I_{2},...I_{n} \rightarrow I_{n},I_{n-1},...I_{1} \end{math}}\\

\resizebox{1\hsize}{!}{\begin{math} Out_{x\_\text{swapped}}= F\left ( \sum_{i=1}^{m}W_{H_iOut_{x}} \cdot F \left ( \sum_{j=1}^{\frac{n}{2}}W_{I_jH_i} \cdot I_{n+1-j} - W_{I_jH_i} \cdot I_j\right )\right )\end{math}}\\

\resizebox{1\hsize}{!}{\begin{math}\therefore Out_{x\_\text{swapped}}= F\left ( \sum_{i=1}^{m}W_{H_iOut_{x}} \cdot F \left ( -\sum_{j=1}^{\frac{n}{2}}W_{I_jH_i} \cdot I_j - W_{I_jH_i} \cdot I_{n+1-j}\right )\right )\end{math}}\\

\resizebox{0.5\hsize}{!}{\begin{math}\because F(x){\text{\;is\;odd\;as\;the\;case\;with\;Eq.\;\ref{eq:sigmoid}}} \end{math}}\\

\resizebox{0.25\hsize}{!}{\begin{math}\therefore F(-x)= -F(x) \end{math}}\\

\resizebox{1\hsize}{!}{\begin{math}\therefore Out_{x\_\text{swapped}}= F\left ( \sum_{i=1}^{m}W_{H_iOut_{x}} \cdot -F \left ( \sum_{j=1}^{\frac{n}{2}}W_{I_jH_i} \cdot I_j - W_{I_jH_i} \cdot I_{n+1-j}\right )\right )\end{math}}

\begin{equation}
\label{eq:eq2}
\resizebox{.91\hsize}{!}{\begin{math}\therefore Out_{x\_\text{swapped}}= -F\left ( \sum_{i=1}^{m}W_{H_iOut_{x}} \cdot F \left ( \sum_{j=1}^{\frac{n}{2}}W_{I_jH_i} \cdot I_j - W_{I_jH_i} \cdot I_{n+1-j}\right )\right )\end{math}}
\end{equation}

\begin{equation}
\label{eq:eq3}
\resizebox{.35\hsize}{!}{\begin{math}\therefore Out_{x\_\text{swapped}}= -Out_{x}\end{math}}
\end{equation}

So, according to the above analysis, in equations (\ref{eq:eq1}), (\ref{eq:eq2}), and (\ref{eq:eq3}), it is evident that when the input values are swapped,  due to the constrained weights, the output will be inverted as well, allowing for a robust change in direction improving obstacle avoidance. Therefore, experiments are conducted to evaluate this result and test its impact on the control system's performance. Furthermore, the equation (\ref{eq:eq1}) formula is recursive; where outputs of one layer are used as input to the next layer, it can be applied to multiple hidden layers networks (deep networks). 

\subsection{Genetic algorithms framework}
\begin{figure}[htbp]
\centerline{\includegraphics[scale=0.3]{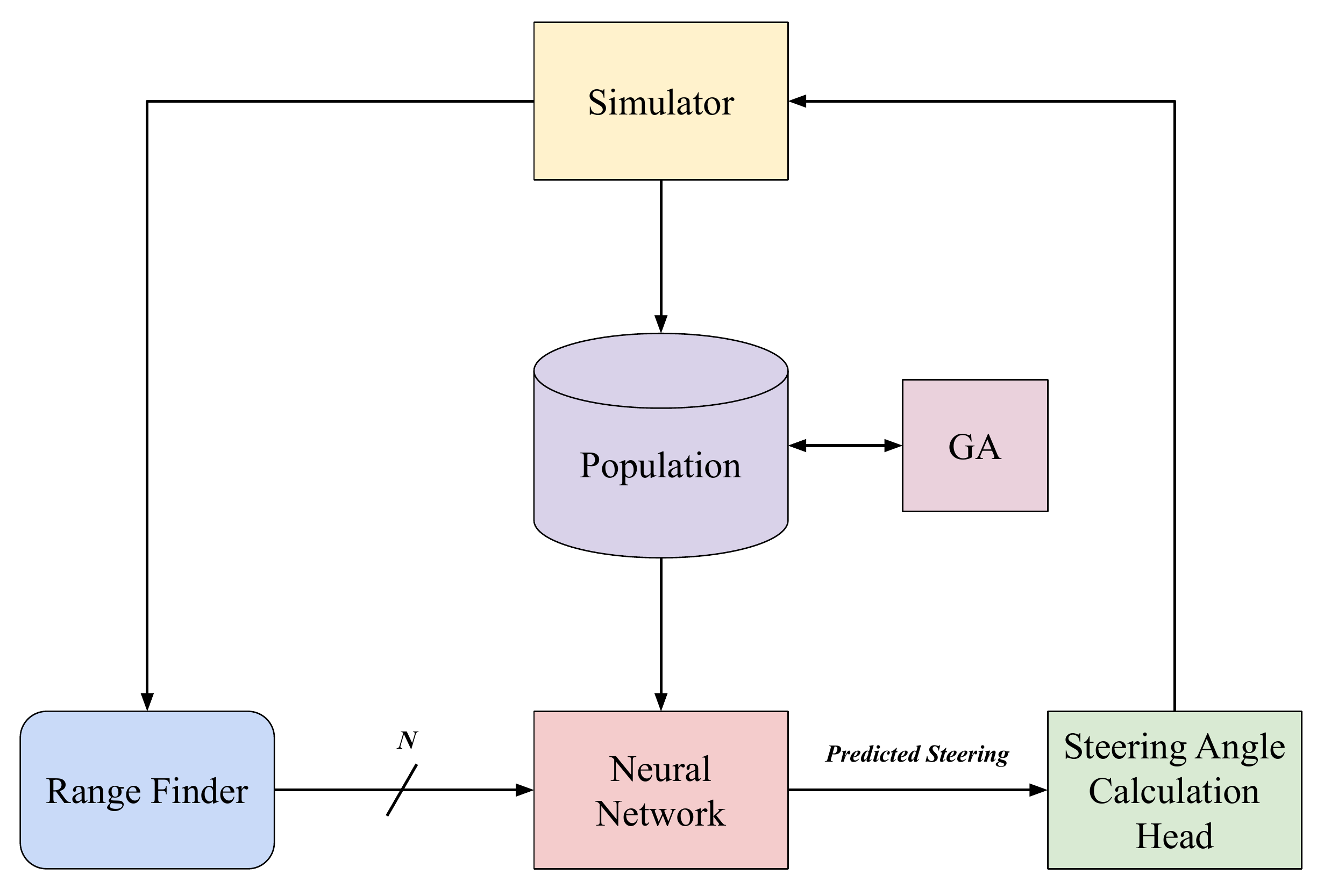}}
\caption{System overview during the training phase. GA is used to generate the population. The population is composed of chromosomes representing the weights of the Neural Network. Each chromosome is used to initialize the weights of the network. Then the network predicts the maneuvering of the vehicle. The maneuver results are fed to the simulator to calculate the fitness of the used chromosome using the genetic algorithm.}
\label{fig3}
\end{figure}

\begin{figure}[htbp]
\centerline{\includegraphics[scale=0.3]{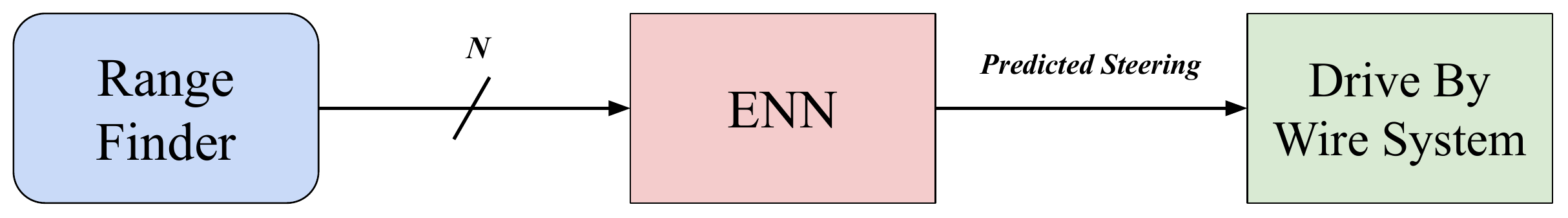}}
\caption{The trained network is used to generate steering commands from a single front-facing proximity sensor.}
\label{fig4}
\end{figure}

We have developed a simulation setup to evaluate the fitness of each chromosome in a generation. The fitness function, Eq. \ref{eq:fitness}, has been chosen to minimize time and maximize the distance covered by the vehicle.

\begin{equation}
\label{eq:fitness}
\resizebox{.35\hsize}{!}{\begin{math}Fitness = \frac{distance^{2}}{time}\end{math}}
\end{equation}

where distance is measured from the track's starting point, and time is measured in simulation ticks. GA is used to pick and breed generations of more intelligent vehicles. The vehicle uses a rangefinder sensor that calculates $\mathbf{N}$ intersections depths with the environment and then feeds these values as inputs to the neural network. The inputs are then passed through a multi-layered neural network and finally to an output layer of $\mathbf{2}$ neurons: steering force to left and right. These forces are used to turn the vehicle by deciding the vehicle's steering angle. Figure \ref{fig3} shows the proposed system overview for our method during the system-training phase.
Once trained, the neural network is able to generate steering commands from the input rangefinder sensor readings. Figure \ref{fig4} shows this configuration.

\section{Simulation Setup}
\subsection{Vehicle model}
A simple 2D bicycle model was used to model the vehicle's longitudinal and lateral motion where $\mathbf{T}$ is the vehicle's center of gravity. The 2D bicycle model assumes a front driving model for the vehicle. The model controls the vehicle driving parameters \begin{math}[x, y, \theta, \delta]\end{math}. Where $\mathbf{x}$, and $\mathbf{y}$ are the current state center of the vehicle, $\mathbf{\theta}$ is the heading angle, and $\mathbf{\delta}$ is the steering angle. This model is effective for controlling small robotic vehicles in normal driving scenarios. At the simulation start, the vehicle is positioned at the start of the track.

\subsection{Sensor model} 
Five different sensor models have been tested, two ideal sensors and three noisy sensors. All the used sensors are based on the point cloud model, where the sensor sensation points are reflected with the surrounding environment and return the distance value of each beam in pixels. 
For the ideal sensors, a basic beam sensor is tested where the number of range beams is configurable, with a constant field of view angle of 180$^{\circ}$. The beams are equally distributed across the field of view. Also, a pseudo camera sensor is tested where the horizontal field of view is 100$^{\circ}$, and the horizontal sight range is 1.5 times the track width in pixels.
For the noisy sensors, a pseudo LIDAR sensor is used, with a configurable number of points in the point cloud with 145$^{\circ}$ field of view angle. A white noise model is applied to the range using the Normal distribution model with one as mean and 0.05 for standard deviation. The noise model affects the accuracy of the distance measured by the sensor for each point in the point cloud. In addition, a pseudo-long-range RADAR is used with a configurable number of points in the point cloud and 110$^{\circ}$ field of view angle. The range noise model follows the Normal distribution with one as mean and 0.1 for standard deviation. Also, a pseudo-medium-range RADAR is used with a configurable number of points in the point cloud and 160$^{\circ}$ field of view angle. The range noise model is used following the Normal distribution with zero as mean and 0.15 for standard deviation, which means that for this sensor, the noise can result in range points wrongly sensed at zero distance.

\section{Experimental Work}
\begin{figure}[htbp]
\centerline{\includegraphics[scale=0.25]{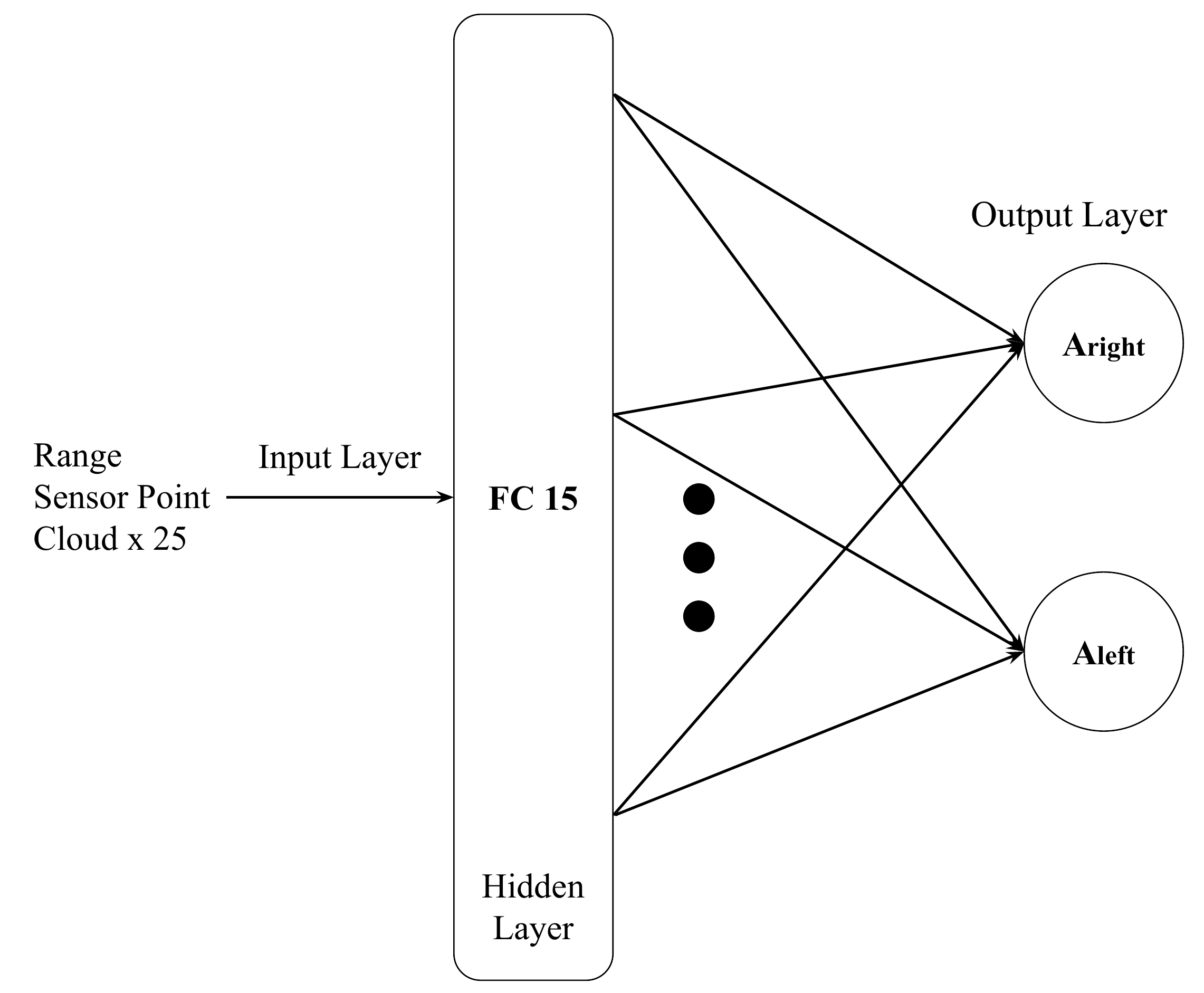}}
\caption{The Symmetric Neural Network is based on Feed Forward Neural Network Architecture used to drive the test vehicle using the front rangefinder sensor readings as input. The architecture is composed of 3 Layers. First Layer is the input layer connected to the range finder sensor raw data. The input layer is connected to a fully connected layer. The Fully connected layer is connected to the final output neurons.}
\label{fig7}
\end{figure}

Several experiments are conducted to evaluate the method of the symmetric neural networks, its impact on the performance of the required tasks, and its robustness to the noise. In addition, the learning approach has been tested to study its impact on our method. The main objective is to inspect the feasibility of a reactive collision avoidance system using our proposed method. Initially, an elementary and relatively straightforward experiment is conducted using a simple map and an ideal sensor. The objective of this experiment is to examine the capability of a vehicle to learn the task of self-navigation through a static environment that does not include any obstacles or sensor noise. This task is less challenging than the collision avoidance task because the ego-vehicle does not have to deal with any uncertainty. This experiment evaluates several neural network models to select a model for the rest of the experiments. The initial experiment has helped us decide on an appropriate model: a three-layer ANN with sigmoid activation function, Eq. \ref{eq:sigmoid} for all neurons. Figure \ref{fig7} shows the network architecture. 

It is noted that the experiment's results do not change if the number of layers is changed, but sometimes the same results are obtained faster. The higher the number of hidden layers, the better representation of the data the network can achieve. Nevertheless, this leads to a more complex optimization problem that is harder and slower for GA to solve. Our GA uses a population of 200 chromosomes where mutation probability is 0.1, crossover probability is one, and the crossover site follows a normal distribution with a mean of 0.95 and a standard deviation of 0.05. The selection is based on tournaments of size ten candidates, and children of next generations always replace their parents. The fitness function is chosen to be the square of the distance from the start of the track to the vehicle's first collision point divided by the vehicle lifetime.
The experiments are done on challenging maps with static obstacles, and specific turn points to challenge the vehicle into changing the speed and the steering angle dramatically while driving. 
The experimental work results encourage and validate the effectiveness of the proposed method.

\subsection{Constrained vs. unconstrained network model}
\begin{figure}[htbp]
\centerline{\includegraphics[scale=0.35]{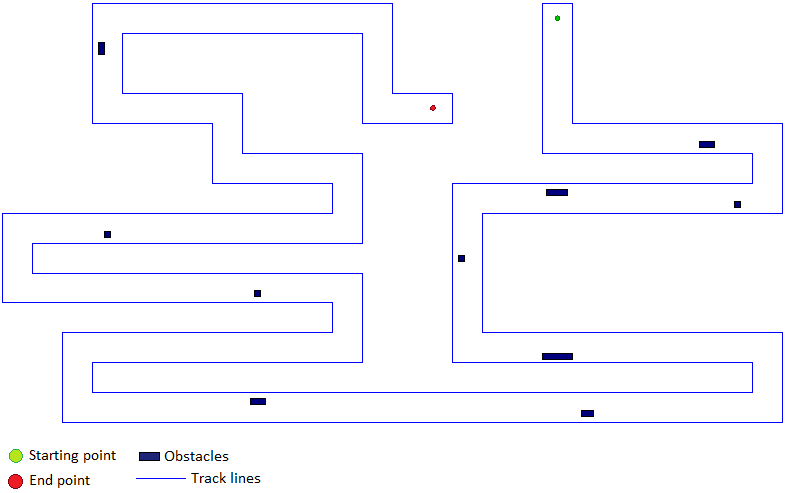}}
\caption{example of tracks used for experiments. The track is composed of horizontal and vertical lines representing the walls and static objects to increase the level of the challenge. The green dot represents the starting position of the vehicle. The red dot represents the destination.}
\label{fig6}
\end{figure}

This experiment aims to compare the effectiveness of the proposed constrained network model vs. the conventional unconstrained network model (ANN) on the learning behavior of the vehicle. The vehicle should navigate a static environment without colliding with the environment's objects or boundaries. The environment is represented by a track composed of several horizontal and vertical edges and static obstacles represented by rectangles of varying dimensions. Figure \ref{fig6} shows one example of several experimental tracks used. The experiment has been done using the ideal simple sensor and has been done for ten different challenging test tracks.

\begin{figure}[htbp]
\centerline{\includegraphics[scale=0.4]{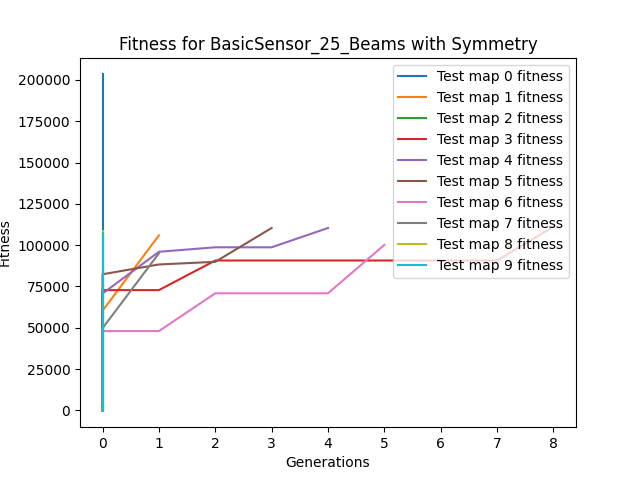}}
\caption{Vehicle navigation performance with a constrained network (symmetric neural network). Front basic sensor with 25 beams is used. In the worst case, the system has required eight generations to reach a winning chromosome that can be used for map navigation, which is expected since the symmetry enforced a shorter chromosome size and so it has simplified the search space of the genetic algorithm, which has reduced the required number of generations to reach a fitting chromosome.}
\label{fig8}
\end{figure}

\begin{figure}[htbp]
\centerline{\includegraphics[scale=0.4]{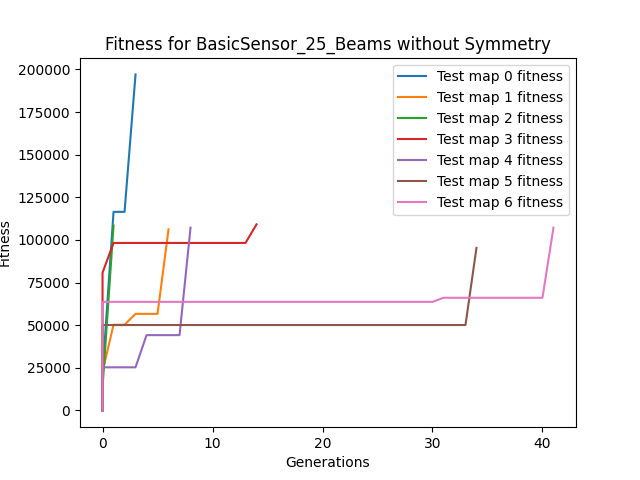}}
\caption{Vehicle navigation performance without constrained network (using traditional artificial neural network). Compared to the symmetric neural networks, using traditional networks the system required a greater number of generations to reach a winning chromosome that can be used for map navigation.}
\label{fig9}
\end{figure}

The experimental results are shown in Figure \ref{fig8}. As illustrated, the vehicle has successfully learned navigation in less than five generations using the constrained symmetrical network model versus more than 40 generations in conventional neural networks. These results are expected since the symmetry enforces robust steering decisions, and also chromosome size required for the symmetric network is shorter due to the symmetry property, so the search space is smaller, and as a result, fewer generations are required to reach a fitting chromosome. In addition, in some maps like maps 7, 8, and 9, the vehicle has failed to successfully navigate to the end of the track even after 100 generations using conventional networks, as shown in Figure \ref{fig9}. Therefore, test maps 7, 8, and 9 have not been presented in the generations' results for the unconstrained networks since the vehicle has failed to navigate them.

\subsection{Generalization across different tracks}
\begin{figure}[htbp]
\centerline{\includegraphics[scale=0.4]{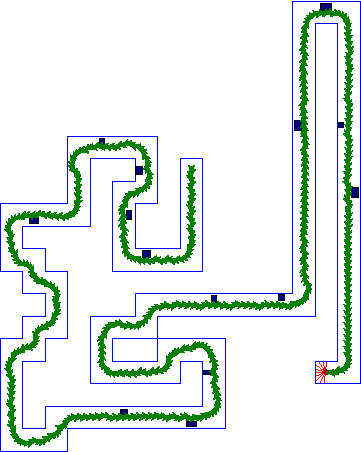}}
\caption{Vehicle maneuvering steps for map six using Front Basic Sensor with 25 Beams.}
\label{fig16}
\end{figure}

\begin{figure}[htbp]
\centerline{\includegraphics[scale=0.4]{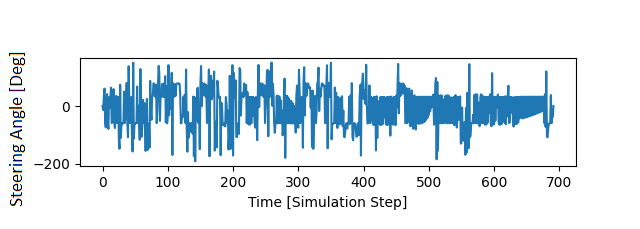}}
\caption{Driving steering angles based on the weights chromosome for map six using Front Basic Sensor with 25 Beams.}
\label{fig17}
\end{figure}

In this experiment, we test the generality of our method across multiple different tracks unseen during training. The training track and results are shown in Figure \ref{fig16}. As illustrated in Figure \ref{fig17} shows the driving parameters used to drive the vehicle through the track. The vehicle can generate steering angles that can accurately maneuver the tracks in challenging situations, like corners, and near obstacles, as illustrated in Figure \ref{fig16}.

\begin{figure}[htbp]
\centerline{\includegraphics[scale=0.4]{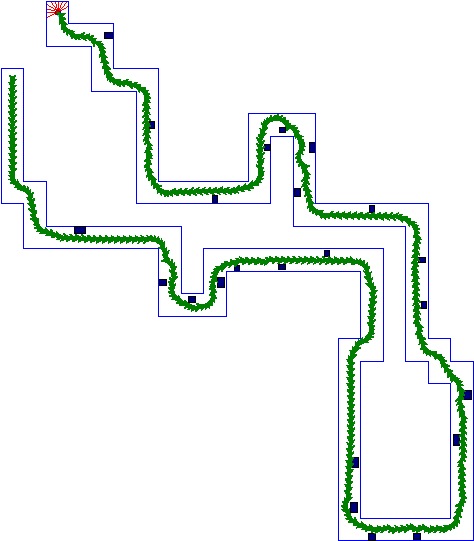}}
\caption{Vehicle maneuvering steps for map seven using Front Basic Sensor with 25 Beams.}
\label{fig18}
\end{figure}

\begin{figure}[htbp]
\centerline{\includegraphics[scale=0.4]{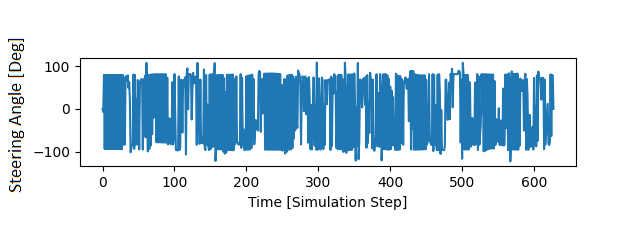}}
\caption{Driving steering angles based on the weights chromosome for map seven using Front Basic Sensor with 25 Beams.}
\label{fig19}
\end{figure}

\begin{figure}[htbp]
\centerline{\includegraphics[scale=0.4]{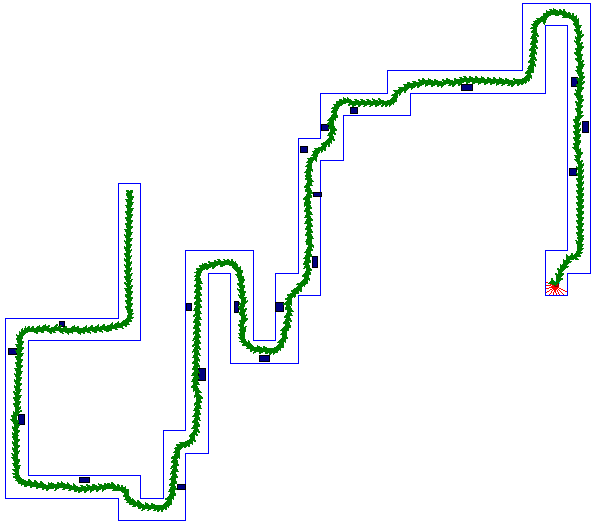}}
\caption{Vehicle maneuvering steps for map eight using Front Basic Sensor with 25 Beams.}
\label{fig20}
\end{figure}

\begin{figure}[htbp]
\centerline{\includegraphics[scale=0.4]{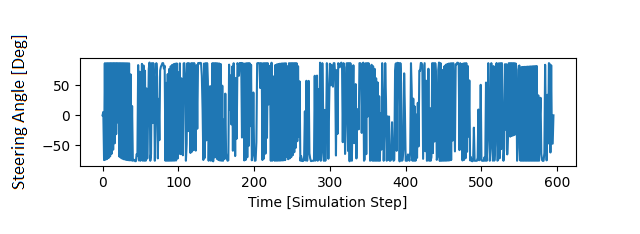}}
\caption{Driving steering angles based on the weights chromosome for map eight using Front Basic Sensor with 25 Beams.}
\label{fig21}
\end{figure}

\begin{figure}[htbp]
\centerline{\includegraphics[scale=0.4]{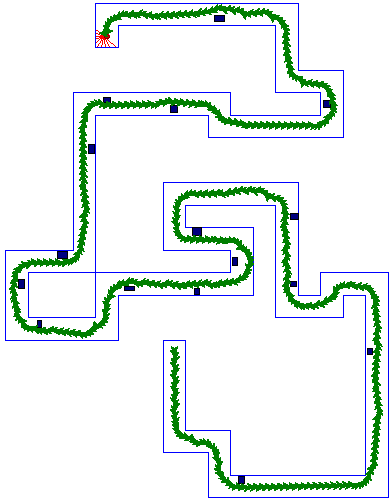}}
\caption{Vehicle maneuvering steps for map nine using Front Basic Sensor with 25 Beams.}
\label{fig22}
\end{figure}

\begin{figure}[htbp]
\centerline{\includegraphics[scale=0.4]{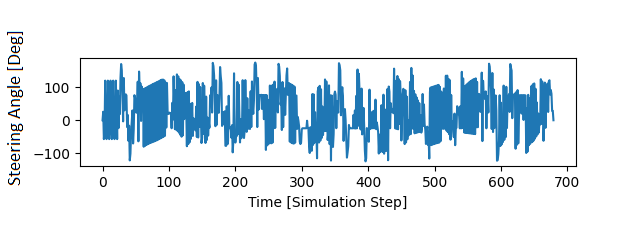}}
\caption{Driving steering angles based on the weights chromosome for map nine using Front Basic Sensor with 25 Beams.}
\label{fig23}
\end{figure}

The same chromosome has been then tested in 3 different randomly generated tracks and results are shown in Figure \ref{fig18}, Figure \ref{fig19}, Figure \ref{fig20}, Figure \ref{fig21}, Figure \ref{fig22} and Figure \ref{fig23}.

The results show that our method generalizes well to the random tracks unseen during training. Furthermore, the vehicle successfully has navigated the three tracks without collision while maintaining the desired steering control behavior. The impact of the symmetric neural network is evident from the stability of the steering angle. Furthermore, the network has been able to generalize to different scenarios using the same chromosome. 

\subsection{Generalization versus different sensor resolution}
\begin{figure}[htbp]
\centerline{\includegraphics[scale=0.3]{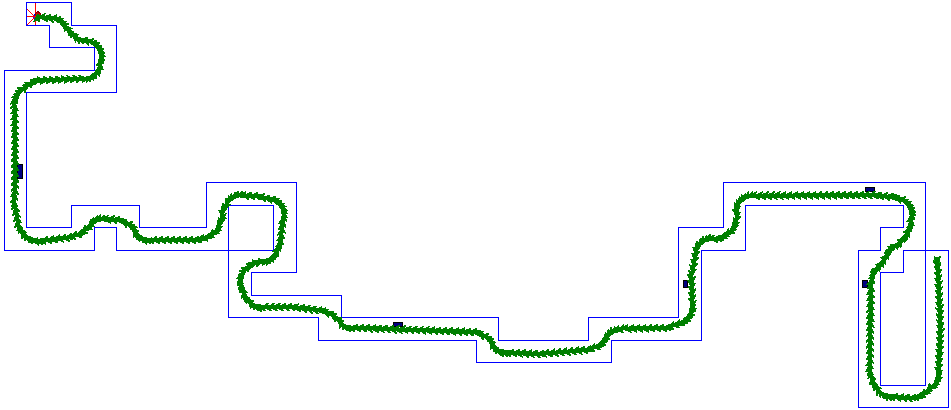}}
\caption{The challenging training track used for chromosome testing.}
\label{fig30}
\end{figure}

This experiment aims to study the effect of the input rangefinder sensor resolution on the learning process using symmetric neural networks. The vehicle has been set to navigate the track shown in Figure \ref{fig30} while varying the number of sensor beams each time. The angle between adjacent beams is equal. The sensor's horizontal range is chosen 180° in our experiment.

\begin{figure}[htbp]
\centerline{\includegraphics[scale=0.4]{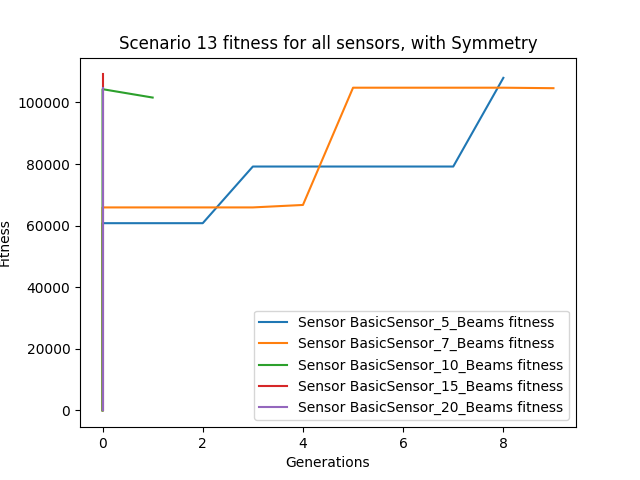}}
\caption{Basic sensors with different beam settings. The basic sensor is used for navigating the challenging track in Figure \ref{fig30}. The basic sensor is used with five beams, seven beams, ten beams, 15 beams, and 20 beams. As illustrated, using 5 and 7 beams, the system has required many generations to reach a chromosome that can navigate the map. However, as the number of beams increases, fewer generations are required to reach a winning chromosome.}
\label{fig27}
\end{figure}

As shown in Figure \ref{fig27}, using a sensor with seven or fewer beams required longer training time due to insufficient input data to the network, as a smaller number of beams will not be able to describe the obstacles along the track adequately. On the other hand, a more significant number of beams (i.e., 15 or more) allows the vehicle to learn and reach acceptable fitness.

\subsection{Generalization versus sensor noise}
\begin{figure}[htbp]
\centerline{\includegraphics[scale=0.4]{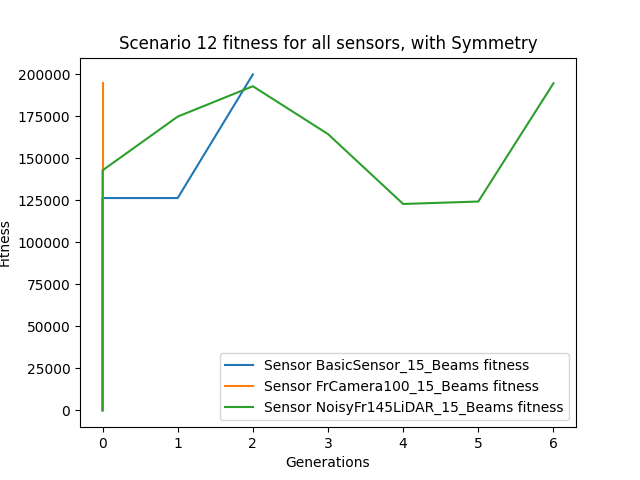}}
\caption{Basic Sensor, with 15 beams, Front Camera with 15 pixels, and Front Noisy LIDAR with 15 beams are tested. The testing is done using symmetric neural networks. As illustrated, the Basic sensor and front camera can converge in fewer generations, while the LIDAR, due to noise, has required more generations. However, for the Long Range Noisy RADAR and Medium Range Noisy RADAR, they have not converged due to the very high noise level.}
\label{fig26}
\end{figure}

\begin{figure}[htbp]
\centerline{\includegraphics[scale=0.4]{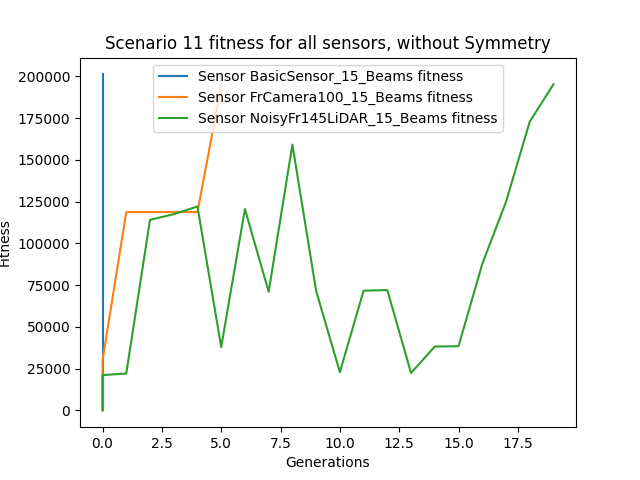}}
\caption{When the same sensors have been tested using traditional neural networks, it has been very complex for them to converge, as illustrated, they have required a larger number of generations. The same results have been observed for the Long Range and Medium Range Noisy RADAR.}
\label{fig25}
\end{figure}

\begin{figure}[htbp]
\centerline{\includegraphics[scale=0.3]{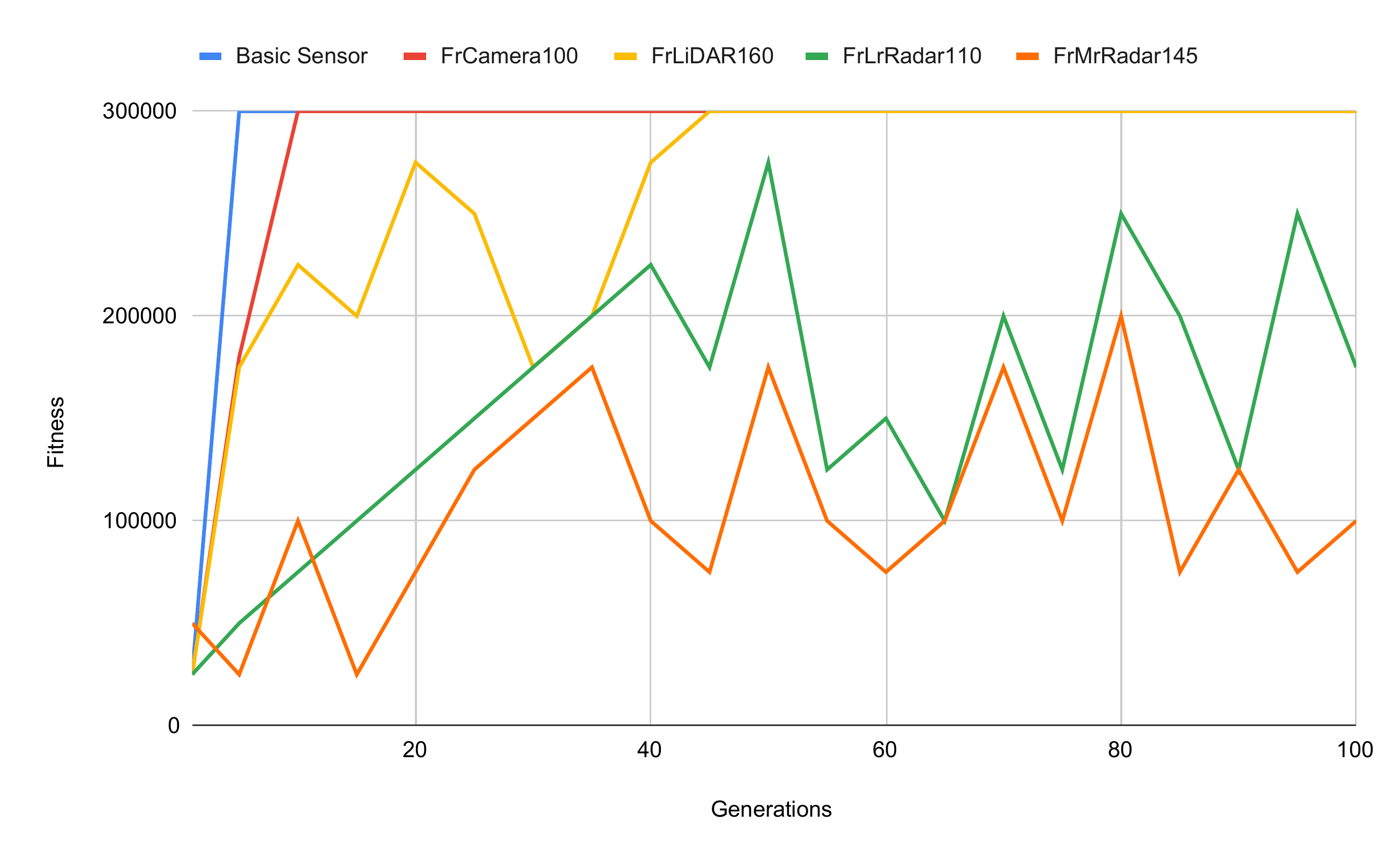}}
\caption{Different sensors performance and learning state. The diagram describes the relationship between the maximum fitness reached by each sensor and the number of generations required to reach such fitness. As illustrated, as the noise level increases, the ability of the sensor to converge to a high fitness chromosome decreases. The oscillations in the fitness between the generations are due to noise. The random noise has caused changes in fitness from one generation to another.}
\label{fig32}
\end{figure}

\begin{figure}[htbp]
\centerline{\includegraphics[scale=0.4]{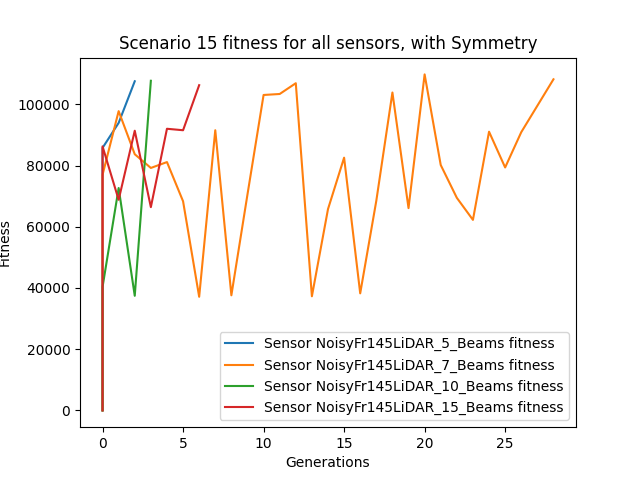}}
\caption{Degradation of the fitness of the pseudo LIDAR sensor. The Front Noisy LIDAR has a noise model of 0.05 standard deviation. At a higher number of range points, the LIDAR can converge to the fitting chromosome.}
\label{fig29}
\end{figure}

The objective of this experiment is to test the tolerance of our method to sensor noise. We have conducted several test runs in the track shown in Figure \ref{fig6} while using different sensor models. The constrained and non-constrained neural network models have been tested. 
The observation is that with using the constrained model, as in Figure \ref{fig26}, symmetric neural networks have required a fewer number of generations compared to unconstrained neural networks as in Figure \ref{fig25}. However, as for the pseudo-Long-range noisy RADAR and pseudo-medium-range noisy RADAR, as they have been tested as well, there have not been any possible results since the noise has been extremely high such that there has been no fitting chromosome found. It has been observed that as the noise increases, the chromosome's fitness has degraded as in Figure \ref{fig32}. Another experiment has been conducted where the same fitting chromosome has been used to drive the vehicle using the pseudo LIDAR sensor with a different number of point cloud points. Figure \ref{fig29} shows that as the size of the point cloud increases, the symmetric neural network can overcome the noise in the pseudo LIDAR and perform adequately. The impact of the noise is evident from the oscillations of the fitness values with each generation. The noise has caused the fitness to be unstable from one generation to another. The fitness trend has been oscillating, unlike in the ideal sensors, where the fitness trend is rising from one generation to another.

\subsection{Chromosome Selection Algorithm}
\begin{figure}[htbp]
\centerline{\includegraphics[scale=0.4]{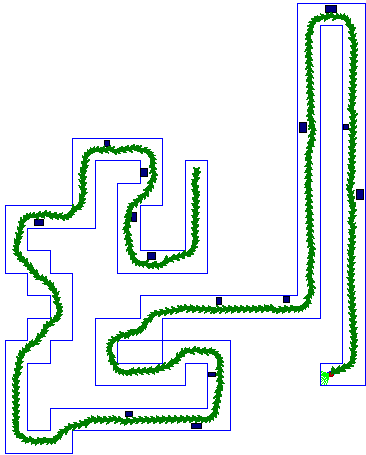}}
\caption{Vehicle maneuvering steps for map 6 using Front Camera with 25 points.}
\label{fig11}
\end{figure}

\begin{figure}[htbp]
\centerline{\includegraphics[scale=0.4]{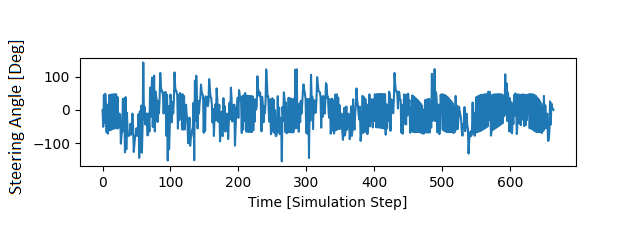}}
\caption{Driving steering angles based on the weights chromosome for map 6 using Front Camera with 25 Beams.}
\label{fig13}
\end{figure}

\begin{figure}[htbp]
\centerline{\includegraphics[scale=0.42]{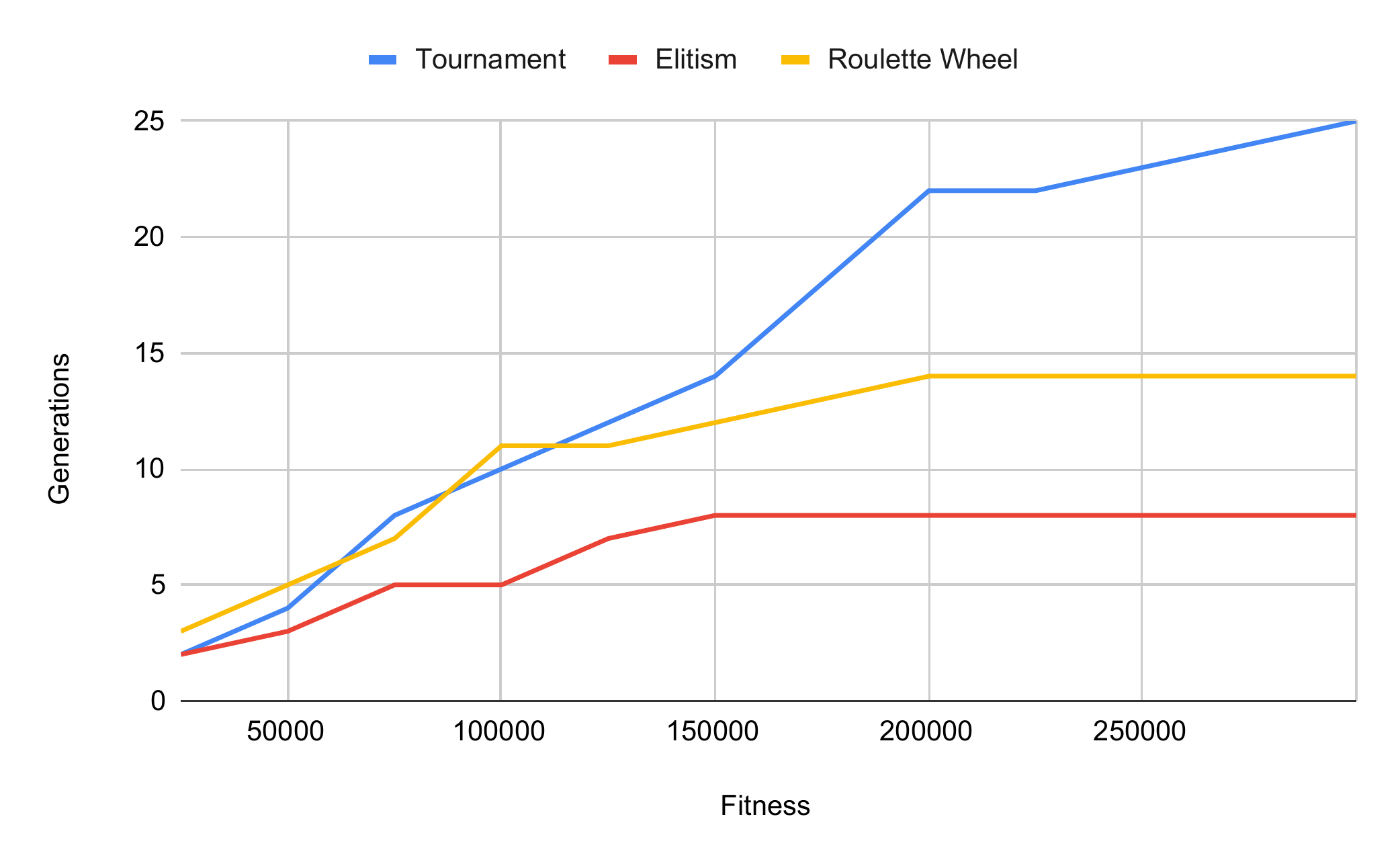}}
\caption{impact of selection method on the number of generations required for finding a fitting chromosome. The diagram compares the number of the required generations required to reach the maximum fitness for the different selection methods. The lower the number of generations, the better the method's performance. As illustrated, Elitism has reached the lowest number of generations to converge to a fitting chromosome.}
\label{fig14}
\end{figure}

Several chromosome selection approaches are tested to train the symmetric neural networks in this experiment. The experiment aims to study which genetic selection approach would suit symmetric neural networks training. The first approach, the tournament approach, where the most ten fitting chromosomes are selected for crossover and mutation to enhance their features, replacing the parents, and the rest of the population is populated from a new random set.
The second approach is the elitism approach, where the ten most fitting chromosomes are selected for crossover and mutation, but rather than replacing them with the offspring, they are kept for the next generation. Therefore, the new generation is composed of the selected parents + the offspring + the newly populated chromosomes.
The third approach is the Roulette Wheel selection, as the higher the parent's fitness, the higher the probability of being selected for breeding. In this method, the algorithm was not constrained to selecting the top 10 candidates, but the chromosomes were sorted and mated on binary bases, where every two highly fitting chromosomes were mated together to generate two new offspring. No new randomly generated chromosomes are required in this approach.
Figure \ref{fig11} shows the performance of the vehicle while using the pseudo camera range sensor to drive through a challenging track. Figure \ref{fig13} shows the driving parameters generated by the symmetric neural network model to control the vehicle. Finally, figure \ref{fig14} shows the comparison between the different selection approaches used to find the most fitting chromosome—the lower the number of the required generations, the better the approach. As illustrated, the elitism approach has given a better performance.

\section{Conclusions}
In this paper, we have proposed a new method of symmetric neural networks based on evolutionary techniques. It has validated the vehicle reactive collision avoidance method with accurate and noisy sensors. Extensive experiments of varying conditions and objectives are conducted to evaluate the proposed method. The results demonstrated in the paper reflect the potential for our proposed method. The vehicle learns to drive collision freely in a static environment. Promising progress is achieved in developing general collision avoidance behavior. The symmetric property of the network has reduced the size of the chromosome and reduced the number of the required generations to find a fitting chromosome.

\section*{Future Work}
Future work will deploy the experiments in more realistic, dynamic, and complex simulation environments. Also, we plan to upgrade the GA operators to further improve our method's performance. In addition, it is planned to study the evolutionary methods for spiking neural networks and apply the method to real-world small robots.

\end{document}